\documentclass[journal]{IEEEtran}
\ifCLASSINFOpdf
\else
\fi

\usepackage[cmex10]{amsmath}
\usepackage{amssymb}
\usepackage{graphicx}
\usepackage{hyperref}
\usepackage{comment}

\begin{document}

\title{Volcanic Clouds Detection through QCNN and Geostationary Satellite Multispectral Imagery}
%
%
%
    
\author{Federica~Torrisi,             
        Claudia~Corradino, Alessandro~Grilli, Tommaso~Catuogno, Mattia~Verducci, Elisabetta~Paladino, Luigi~Giannelli, Alessandro~Sebastianelli
\thanks{F. Torrisi and C. Corradino are with Istituto Nazionale di Geofisica e Vulcanologia (INGV), Sezione Osservatorio Etneo (e-mail: federica.torrisi@ingv.it; corradino.claudia@ingv.it). T. Catuogno, A. Grilli and M. Verducci are with Thales Alenia Space Italia (e-mail: tommaso.catuogno@thalesaleniaspace.com; alessandro.grilli-somministrato@thalesaleniaspace.com; mattia.verducci@thalesaleniaspace.com). A. Sebastianelli is with Euro-Mediterranean Center on Climate Change, REMHI, Caserta, Italy (e-mail: alessandro.sebastianelli@cmcc.it). L. Giannelli and E. Paladino are with Dipartimento di Fisica e Astronomia “Ettore Majorana,” Università di Catania, Catania, Italy and with INFN, sezione di Catania, Catania, Italy (e-mail: luigi.giannelli@dfa.unict.it; elisabetta.paladino@dfa.unict.it ) }
\thanks{Manuscript received *; revised *.}
}

\maketitle

\begin{abstract}
    Recent advances in quantum computing are opening new possibilities for Earth Observation (EO) data analysis. Quantum machine learning (QML) approaches offer novel ways to process information by exploiting quantum phenomena such as superposition and entanglement. These capabilities have motivated the exploration of whether quantum-enhanced models can address long-standing challenges in satellite remote sensing, where complex spectral and spatial signals often require sophisticated feature extraction. Among various fields of application, EO data allow the global monitoring of volcanic clouds and are crucial for aviation safety, hazard assessment, real-time eruption response, and evaluation of volcanic impacts on climate. Yet accurate detection of volcanic clouds remains difficult due to their similarity with meteorological clouds, the variability of eruption signatures, and the coarse spectral sampling of geostationary sensors. In this work, the potential of hybrid quantum convolutional neural networks (QCNNs) for the classification of satellite images containing volcanic clouds was investigated. These architectures integrate quantum computational layers into a classical convolutional framework. Two QCNN variants (with 2 and 4 qubits) have been considered to evaluate their ability to classify a dataset of SEVIRI images, including scenes with volcanic clouds (composed of ash, $SO_2$, or mixed components) as well as non-volcanic backgrounds. Finally, the performance of the hybrid QCNN models was compared with that of purely classical architectures.
\end{abstract}

\begin{IEEEkeywords}
    Earth observation, Quantum Machine Learning, Volcanic Clouds, Parameterized Quantum Circuit
\end{IEEEkeywords}

%
\IEEEpeerreviewmaketitle

\section{Introduction}
%
%
%
%

\IEEEPARstart{I}{n} recent years, the rapid evolution of quantum computing (QC) has unlocked new frontiers for data analysis across diverse scientific disciplines. By leveraging the principles of quantum mechanics, specifically superposition and entanglement, QC introduces a computational paradigm capable of processing complex information in ways that transcend the binary limitations of classical architectures \cite{kaye2006introduction} 
This shift is particularly significant for fields burdened by high-dimensional data, where quantum algorithms are increasingly demonstrating the potential to tackle problems previously considered computationally intractable.
A primary beneficiary of this revolutions is the field of Earth Observation (EO), which has entered an era of Big Data, since the volume of information generated by modern remote sensing missions has reached unprecedented scales \cite{sudmanns2020big}. 
Characterized by vast differences in spatial, spectral, radiometric, and temporal resolutions, this heterogeneous data enhances global change monitoring while creating a severe processing bottleneck.
Artificial intelligence (AI) has become the primary tool for managing high-resolution EO Big Data, profoundly transforming the field through its ability to extract meaningful insights from expanding datasets \cite{tuia2024artificial}.
However, classical machine learning (ML) has not fully resolved the computational challenges. Large-scale ML can be time-consuming and energy-intensive due to inherent limitations related to data management, processing, and efficient resource utilization \cite{wang2020survey}.
To address these challenges, integrating QC into the EO workflow is becoming increasingly important.  
Quantum-enhanced approaches offer superior speed and efficiency for these computationally intensive tasks, enabling remote sensing big data to be processed far faster than with classical systems \cite{sebastianelli2026quantum}.
The growing interest in implementing ML methodologies based on QC has led to the emergence of a new research field known as Quantum Machine Learning (QML) \cite{chang2025primer}. 
Preliminary studies suggest that quantum-enhanced models can explore vast, complex solution spaces more efficiently than classical counterparts, potentially offering faster convergence, superior generalization, and enhanced performance on large-scale datasets \cite{zaman2023survey}.
Research in this area has increasingly focused on hybrid quantum-classical architectures, reflecting the current limitations of quantum hardware, such as limited number of qubits and coherence time. Although these developments are promising, they only partially address the broader challenges of deploying quantum-enhanced models in EO applications \cite{miroszewski2023quantum}.
Despite the field’s early stage, it is driven by the remarkable potential of quantum technologies to transform EO workflows, offering new opportunities for handling complex datasets and performing computations that are beyond the reach of classical approaches.

Among various fields of application, EO data allow the global monitoring of volcanic phenomena, like the dispersion of volcanic clouds, crucial for aviation safety, hazard assessment, real-time eruption response, and evaluation of volcanic impacts on climate. 
A good candidate for the purpose of detecting a volcanic cloud is the Spinning Enhanced Visible and Infrared Imager (SEVIRI) sensor on board the Meteosat Second Generation (MSG) geostationary satellites, which has a good spectral resolution in the infrared region and a high-temporal resolution.
Yet accurate and efficient classification remains challenging due to the variety of eruption signatures and the coarse spectral sampling of geostationary sensors. Furthermore, because explosive eruptions are rare compared to periods of volcanic quiescence, available training datasets remain small, limiting classical ML model performance.
With the idea of providing a solution for those challenges, this work investigates the potential of hybrid Quantum Convolutional Neural Networks (QCNNs) for classifying satellite images containing volcanic clouds. These architectures integrate quantum computational layers into a classical convolutional framework to leverage the advantages of both paradigms. They have already demonstrated superior feature representation when applied to Sentinel-2 MSI imagery for global-scale detection of volcanic thermal activity \cite{corradino2026hybrid}.
Using the SEVIRI dataset, this study analyzes scenes containing volcanic clouds (composed of ash, $SO_2$, or mixed components) alongside non-volcanic backgrounds, such as clear skies and meteorological clouds. 
Two QCNN variants (2-qubits and 4-qubits), have been considered to evaluate the trade-off between classification accuracy and computational cost. Finally, these hybrid models were benchmarked against purely classical architectures. 


\section{Hybrid QCNN architecture}

A hybrid QCNN was developed for scene classification, with the specific aim of detecting volcanic cloud in SEVIRI satellite imagery. The initial version of this hybrid QCNN was introduced in \cite{sebastianelli2021circuit} for Land Use and Land Cover (LULC) classification. Since then, the approach has been progressively improved and further advanced over time \cite{sebastianelli2023quantum}. 

\begin{figure}[!t]
    \centering
    \includegraphics[width=2.5in]{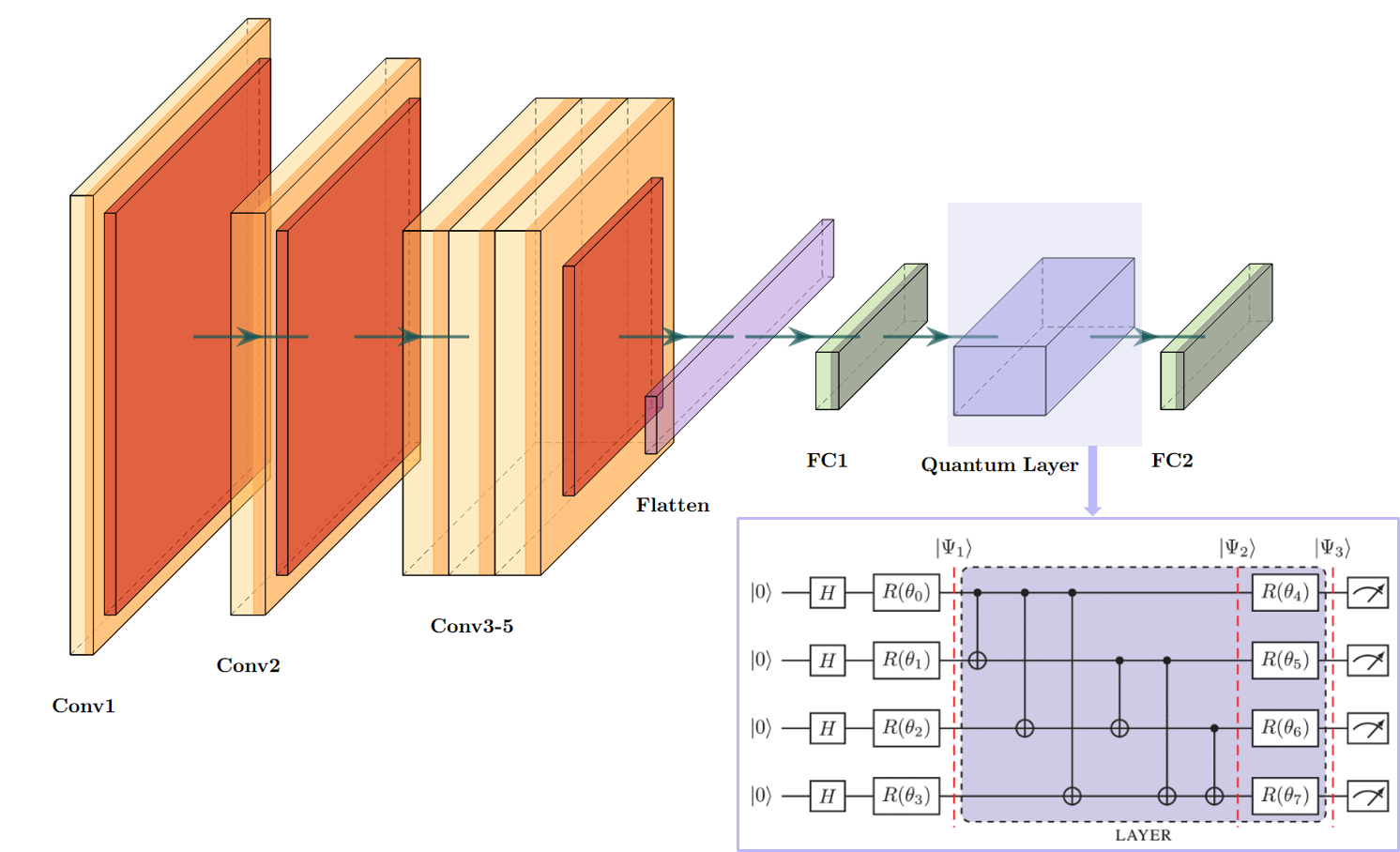}
    \caption{Hybrid Quantum Convolutional Neural Network (QCNN). 
    The blue box indicate the Quantum Circuit  (Real Amplitude). Image adapted from \cite{sebastianelli2021circuit}}
    \label{fig:Figure1_Architecture}
\end{figure}

The scheme of the hybrid QCNN architecture is shown in Fig.~\ref{fig:Figure1_Architecture}. The model combines a classical convolutional neural network (CNN), derived from AlexNet \cite{krizhevsky2012imagenet}, with a parametrized quantum circuit (PQC). This hybrid design aims to improve performance in complex EO tasks, where traditional CNNs may face limitations \cite{sebastianelli2026quantum}. AlexNet consists of five convolutional layers, three max-pooling layers, and two fully connected (FC) layers followed by a softmax classifier.
The convolutional layers are needed to extract hierarchical spatial features from the input images and reduce their dimensionality. 
In the proposed hybrid QCNN architecture, the quantum layer is inserted between two classical FC layers. The first FC layer projects the feature maps extracted by the CNN into a suitable dimension for encoding into the quantum circuit, while the second maps the quantum output into class probabilities for the final classification.
The quantum layer processes the encoded data using parameterized unitary operations within the Hilbert space to generate non-linear feature representations. Before a quantum circuit can process satellite data, classical feature vectors must be mapped into quantum states through a procedure known as \textit{quantum data encoding}. This operation defines how information is represented in the Hilbert space and directly affects the model's expressive power, hardware efficiency, and resilience to noise \cite{sebastianelli2026quantum}. 
The main quantum encoding strategies include: \textit{angle encoding}, where classical features are mapped to rotation angles of single-qubit gates and which is particularly efficient and robust for near-term devices; \textit{amplitude encoding}, where a normalized data vector is directly embedded into the amplitudes of a quantum state, offering high expressive power and exponential compression but requiring more complex state preparation; \textit{hybrid or physics-informed embeddings}, which integrate domain knowledge such as EO physical parameters to better align the quantum representation with the underlying data. Overall, the choice of encoding strongly influences circuit depth, noise resilience, and model expressiveness, with simpler encodings favoring hardware efficiency and more complex ones improving representational capacity.
After encoding, the data is processed by a PQC, in which gate rotation angles are determined by the classical feature vector, allowing the circuit to act as a nonlinear transformation of the input data. Once the circuit is executed, measurement outcomes are collected and passed to subsequent layers of the hybrid network. 

For the proposed architecture, a Real Amplitudes quantum circuit is employed, as it has demonstrated strong performance in remote sensing classification tasks \cite{sebastianelli2021circuit}. 
The structure of a single-layer, four-qubit configuration is illustrated in Fig.~\ref{fig:Figure1_Architecture}. 
The circuit begins by applying Hadamard gates to all qubits, which places the system into a uniform superposition state. Subsequently, the quantum state undergoes variational evolution through parameterized single-qubit rotations combined with entangling operations. This transformation allows the circuit to explore a high-dimensional Hilbert space and learn complex representations of the input data.
In this work, we implemented two hybrid QCNNs utilizing 2-qubit and 4-qubit RealAmplitude variational ansatzes, respectively, to evaluate the effect of qubit scaling on model performance. Guided by previous studies highlighting the efficiency of low-qubit architectures \cite{sebastianelli2021circuit}, we chose these 2-qubit and 4-qubit configurations to assess how well the system performs in the few-qubit regime.
We are emulating these circuits on classical hardware, so we are not accounting for noise.

\subsection{Implementation Details}

The hybrid QCNN architecture integrates QC into a deep learning framework by combining PyTorch, Qiskit, and Qiskit Machine Learning. PyTorch manages the classical neural network layers and backpropagation process, while Qiskit is used to construct the quantum circuit. The integration between the quantum and classical components is achieved through the TorchConnector, which allows the quantum circuit to behave as a differentiable neural network layer.
The core component is the QuantumLayer, which encodes classical input data into a quantum state using the ZZFeatureMap, applies a trainable variational circuit based on the RealAmplitudes ansatz, and finally measures the quantum state through the EstimatorQNN. The measurements, expressed as expectation values of Pauli-Z observables, are returned as outputs and passed to the classical layers. This design enables end-to-end training where both quantum and classical parameters are optimized together within a single learning pipeline.

\subsection{Training and testing}

The optimal hyperparameters for training the proposed hybrid QCNN were determined using the OPTUNA framework. Several configurations were explored by varying the learning rate, batch size, and maximum number of training epochs. The best-performing setup used a learning rate of $10^{-4}$, a batch size of 16, and a maximum of $200$ epochs, with early stopping applied (patience of $15$ epochs) to reduce overfitting and improve training efficiency. Optimization was performed using the Adaptive Moment Estimation (Adam) optimizer to minimize the Cross-Entropy (CE) loss function, which measures the discrepancy between predicted probability distributions and ground-truth labels. High-performance computing was required due to the complexity of the quantum layers. Consequently, the training was conducted on two \textit{NVIDIA GeForce RTX 4090 GPUs}. This hardware configuration provided a speedup over CPU-based training, which was found to be significantly slower.

Hybrid QCNNs with $2$ and $4$ qubits were trained and evaluated to assess the trade-off between model complexity and classification performance. 
For benchmarking purposes, two classical counterparts were implemented by replacing the quantum circuit with classical neural network structures. In the first classical architecture, the quantum circuit was replaced by a single FC layer: different configurations were then tested by varying the number of nodes in this layer to evaluate how performance scales when then number of parameters changes. In the second classical architecture, the quantum circuit was substituted with a multi-layer perceptron (MLP) featuring sequential FC layers of $256$, $64$, $32$, and $10$ nodes, respectively. To ensure a fair comparison, all classical models were trained under the exact same hyperparameter settings as the hybrid quantum architectures.
Model evaluation was conducted on a test set of unseen images. Performance is reported using confusion matrices and standard metrics, including accuracy (A), precision (P), recall (R), and F1-score (F1). 

\section{Results}

\begin{figure}[t]
    \centering
    \includegraphics[width=0.70\columnwidth]{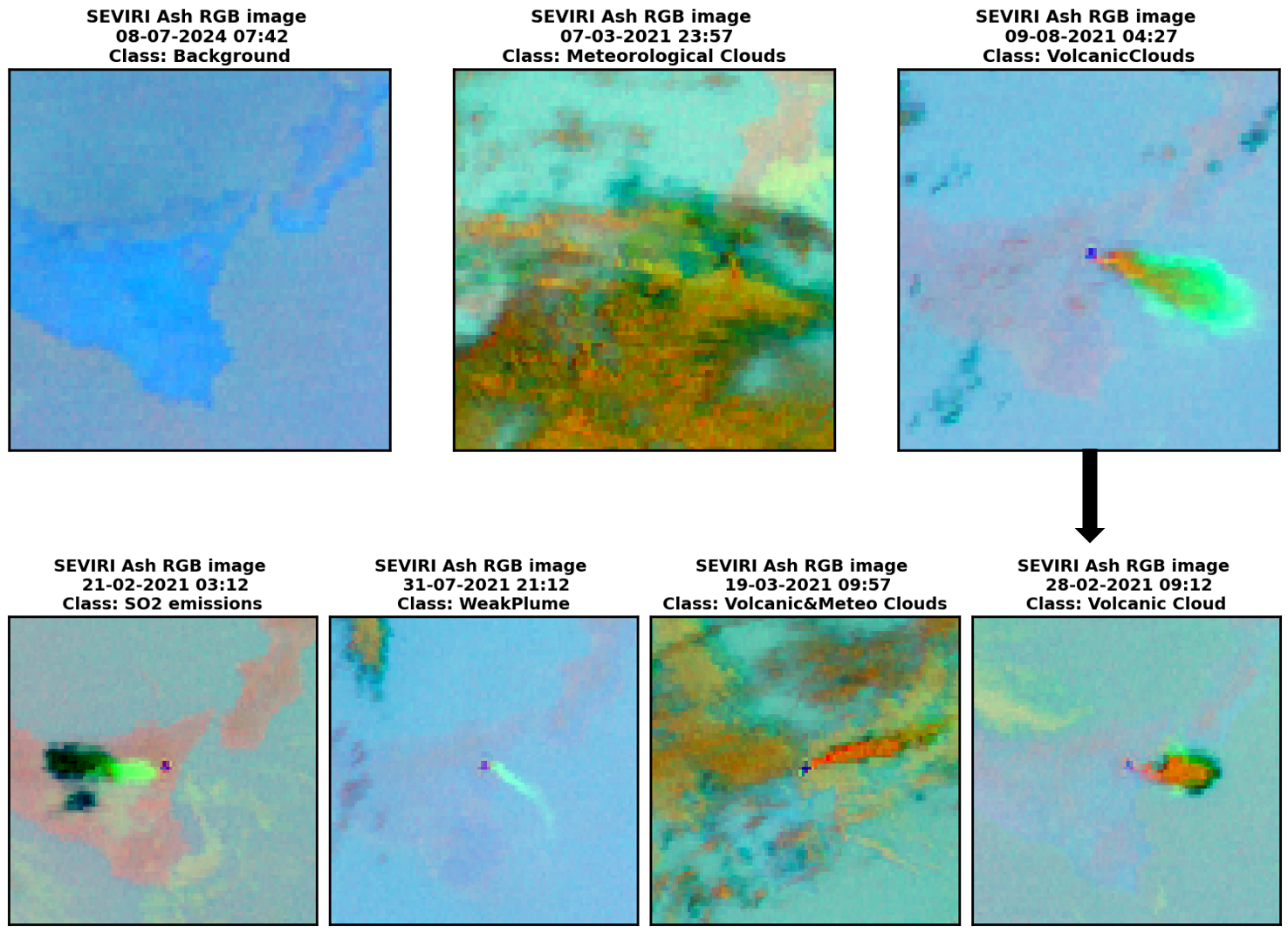}
    \caption{Two-stage classification scheme using SEVIRI Ash RGB composite. Top: 3-class volcanic cloud detection (Background, Meteorological Clouds, Volcanic Clouds). Bottom: 4-class volcanic clouds discrimination ($SO_2$, Weak Plume, Volcanic-Meteo Clouds, Volcanic Cloud with Mixed Components).}
    \label{fig:Figure2_Classes}
\end{figure}

This section presents the results obtained by applying the hybrid QCNN model to the Ash RGB SEVIRI dataset \cite{torrisi2024enhancing}. The input dimensions are $100 \times 100 \times 3$, corresponding to the three-channel RGB composites. Two experimental campaigns were conducted. 
In the first campaign, the SEVIRI dataset was used to perform scene-level classification, with the objective of determining whether an observed scene contained $(1)$ clear sky conditions, $(2)$ meteorological clouds, or $(3)$ volcanic clouds.
In the second campaign, the classification task was refined for cases where volcanic clouds were detected, introducing a more detailed four-class scheme. The classes considered were: $(1)$ $SO_2$ emissions, $(2)$ volcanic clouds, $(3)$ weak plume and volcanic clouds, and $(4)$ meteorological clouds. This second set of experiments aimed to evaluate the model’s ability to distinguish between different types and intensities of volcanic emissions, as well as to discriminate them from non-volcanic cloud formations.
\begin{figure*}[t]
    \centering
    \includegraphics[width=0.9\textwidth]{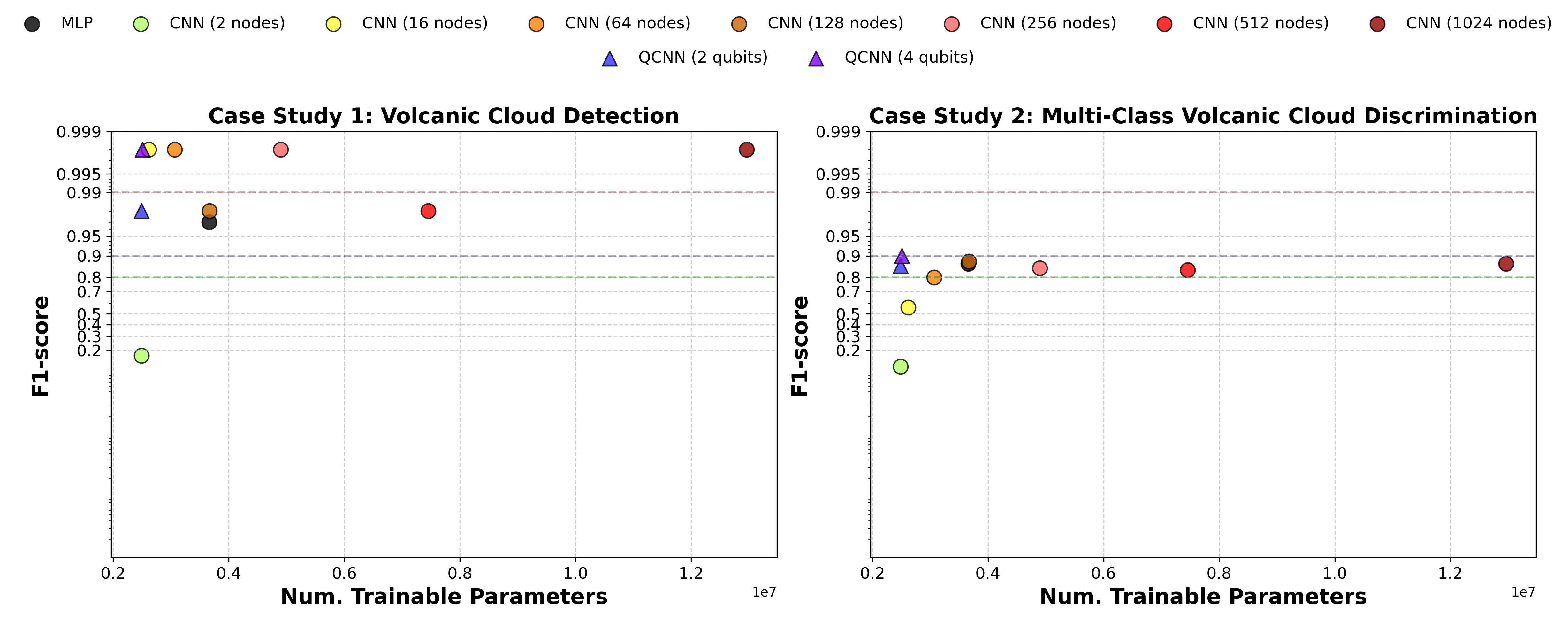}
    \caption{F1-score relative to the number of trainable parameters across classical architectures (MLP, CNN variants) and hybrid models (QCNN 2-qubit and 4-qubit) for Case Study 1 (left) and Case Study 2 (right).}
    \label{fig:Figure3_Results}
\end{figure*}
For the first case study (Case Study I: Volcanic Cloud Detection), a SEVIRI dataset consisting of $791$ Ash RGB images was categorized into three distinct classes: clear sky, meteorological clouds, and volcanic clouds. The primary objective of this classification is to detect the presence of volcanic clouds in any new scene to facilitate a rapid response during volcanic eruptions. The dataset was partitioned into training, validation, and testing sets using a ratio of $70\%$, $15\%$, and $15\%$, respectively. Specifically, this distribution comprises $553$ images for training, $119$ for validation, and $119$ for testing. An example of image for each of the three classes is illustrated in Fig.~\ref{fig:Figure2_Classes}.
For the second case study (Case Study II: Multi-class Volcanic Cloud Discrimination), a SEVIRI dataset of images containing volcanic clouds was employed with the task of classify each of these scene into four distinc classes, according to the type of emission: $SO_2$ emission, Weak Plume, Volcanic Clouds, Mixed Meteorological and Volcanic Clouds with mixed components. The SEVIRI dataset of $764$ Ash RGB images was partitioned into training, validation, and testing sets using a $70/15/15\%$ ratio, resulting in $534$ images for training, $115$ for validation, and $115$ for testing. Representative examples for each of these complex classes are in Fig.~\ref{fig:Figure2_Classes}.

The results obtained from training the hybrid QCNNs (with $2$ and $4$ qubits) and the benchmarking architectures are reported in Fig.~\ref{fig:Figure3_Results}. For the Case Study I (Volcanic Cloud Detection), the hybrid QCNNs with $2$ and $4$ qubits achieved an F1-score of $0.97$ and $1.00$, respectively, demonstrating remarkably high classification accuracy. However, a primary distinction between the two quantum configurations is their computational cost, as the training time for the $4$-qubit model was considerably higher than that of the $2$-qubit version. When compared to their classical counterparts, the hybrid models consistently demonstrated superior accuracy. Classical models with larger FC layers (ranging from $16$ to $1024$ nodes) yielded F1 between $0.95$ and $1.00$. However, their number of trainable parameters scaled from approximately $2.5$ million to $13$ million. This represents an increase from $10^6$ to $10^7$, corresponding to a growth of roughly one order of magnitude. Furthermore, evaluating a classical CNN configured with exactly the same number of parameters as the quantum alternative (i.e., a classical network constrained to just $2$ nodes) resulted in a significantly lower accuracy of $0.12$. 
A similar pattern was observed in the Case Study II (Multi-class Volcanic Cloud Discrimination). However, this task is inherently more challenging because it requires distinguishing among multiple types of volcanic emissions rather than separating volcanic and non-volcanic clouds only. Consequently, the analysis must account for the different spatial structures and spectral signatures associated with each emission type. Consequently, the overall performance is lower, reflecting the increased complexity of the task, but remains robust, with F1-scores consistently above $0.80$. The hybrid QCNN models achieve F1-scores of $0.85$ and $0.90$ for the $2$-qubit and $4$-qubit architectures, respectively. In contrast, none of the classical benchmark models reaches an F1-score of $0.90$, even when using larger architectures with a greater number of trainable parameters. These results suggest the superior generalization capability of the quantum models, particularly in more challenging classification scenarios.

\begin{figure*}[t]
    \centering
    \includegraphics[width=0.85\textwidth]{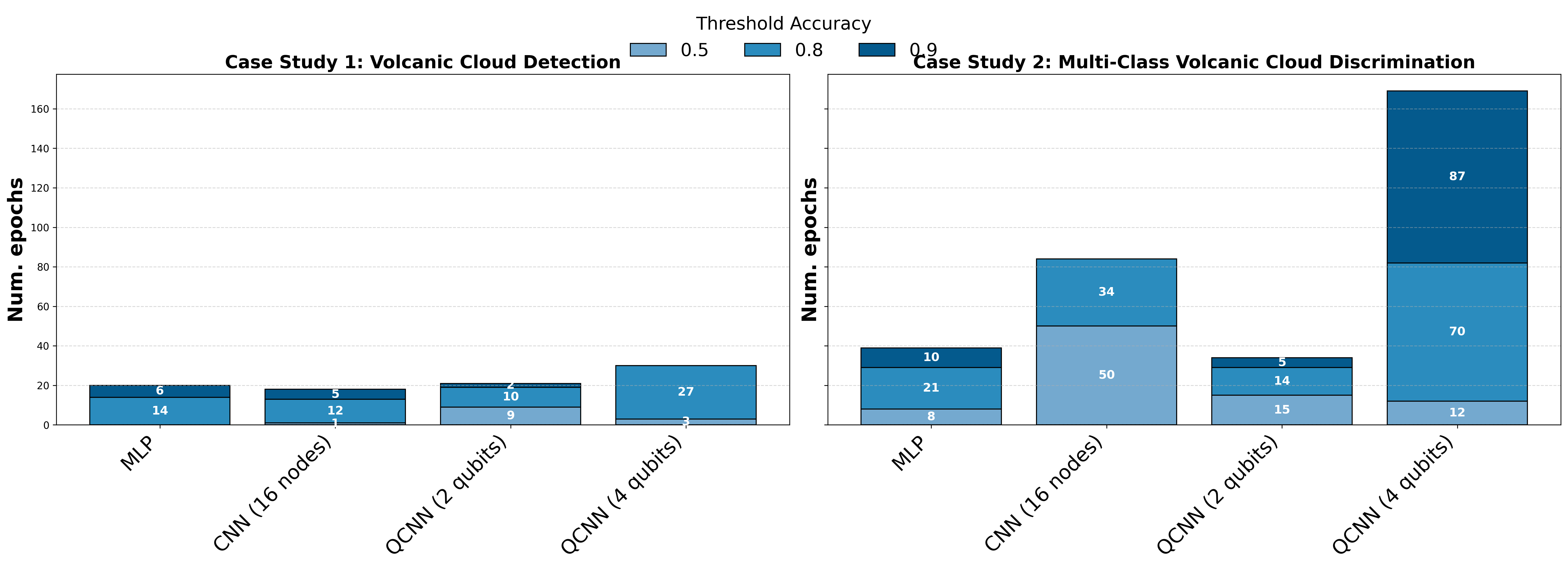}
    \caption{Number of training epochs required by the two hybrid QCNN architectures ($2$-qubit and $4$-qubit) and the two classical benchmark architectures (MLP and CNN with 16 nodes) to reach specific accuracy thresholds ($0.5$, $0.8$, and $0.9$) for Case Study I (left) and Case Study II (right).}
    \label{fig:Figure4_Results}
\end{figure*}

A comparative analysis was performed to measure the training steps required for both hybrid QCNN architectures ($2$-qubit and $4$-qubit) and two classical benchmarks to reach target accuracy thresholds ($0.5$, $0.8$, and $0.9$) across both case studies. In one benchmark, the quantum circuit was replaced by the FC layer with $16$ neurons (CNN with $16$ nodes), while in the other, it was replaced by the MLP (Fig.~\ref{fig:Figure4_Results}).
For the Case Study I (Volcanic Cloud Detection), the classical architectures (MLP and CNN) exhibited rapid convergence, reaching their target accuracy thresholds within fewer than $30$ epochs. In contrast, the hybrid quantum models required substantially longer training. The $2$-qubit QCNN required $113$ epochs to achieve an accuracy of $0.80$. This slower convergence may be attributed to optimization challenges commonly encountered in parameterized quantum circuits, such as barren plateaus and vanishing gradients, particularly for low-dimensional or linearly separable tasks \cite{mcclean2018barren}. Increasing the quantum capacity to a $4$-qubit QCNN improved convergence, reducing the required training to $69$ epochs while enabling the model to reach the $0.90$ accuracy threshold. Nevertheless, it required significantly more training epochs than the classical models, indicating that for relatively simple classification problems, classical architectures provide superior computational efficiency and faster optimization.
When transitioning to the more challenging multi-class classification task, the relative performance of the architectures changed substantially.
The $2$-qubit QCNN emerged as the fastest model to reach the $0.90$ accuracy threshold, requiring only $34$ epochs. This result suggests that even a small quantum feature space may offer enhanced representational power for multi-class decision boundaries, allowing the hybrid model to learn more efficiently than the classical CNN with a $16$-node replacement layer, which required 84 epochs to reach only $0.8$ accuracy.
However, this advantage did not persist as the quantum circuit size increased, since the $4$-qubit QCNN required 169 epochs to converge. As the number of qubits increases, the parameter space grows exponentially, resulting in a more complex optimization landscape with a higher likelihood of barren plateaus and numerous local optima. Consequently, optimization becomes significantly more difficult, with the model requiring 87 additional epochs to improve its accuracy from $0.80$ to $0.90$.
Overall, the results suggest that small quantum circuits can improve learning for more complex tasks, but adding more qubits does not necessarily lead to better performance. In our simulations, the larger quantum circuits are harder to optimize and exhibit slower convergence, removing the potential benefit of the enlarged Hilbert space. These findings indicate that circuit size should be chosen by balancing representational capacity against trainability, rather than by increasing the qubit count alone. Future work will therefore examine how classification accuracy and convergence rate change with the number of qubits, and whether compact circuits can be implemented as resource-efficient modules for on-board data processing.

\section{Conclusions}

In this work, the potential of hybrid QCNNs for classifying satellite images containing volcanic clouds was investigated. The objective was to classify SEVIRI satellite scenes according to the presence of volcanic clouds and the type of volcanic emissions. Two case studies of varying complexity were established. The first focuses on a simpler tri-class discrimination between clear-sky, meteorological clouds, and volcanic plumes.
The second presents a more complex task, i.e. classifying volcanic emission types and intensities, providing a demanding testbed to demonstrate the ability of quantum models in handling high-dimensional decision boundaries.
The results show that hybrid QCNNs can achieve competitive classification performance with significantly fewer trainable parameters than classical models. In both case studies, the $4$-qubit QCNN reached the highest accuracy. However, convergence analysis indicates that increasing the number of qubits generally leads to longer training times, revealing a trade-off between model complexity and optimization efficiency. Notably, even the $2$-qubit QCNN achieved good accuracy, suggesting that compact architectures could be suitable for onboard satellite processing. Overall, these findings highlight hybrid quantum models as a promising parameter-efficient approach for satellite image classification, while also emphasizing the need to improve scalability and training efficiency for larger quantum circuits. Due to its low qubit requirement, this approach is well-suited for onboard processing on EO satellite platforms. Future work will involve testing these models on real quantum processors to evaluate the impact of hardware noise.

\section*{Acknowledgment}
F.T. is supported by the DEMETRA research line within the ROSE (Reinforcement of the Observational Systems of the Earth) infrastructural project of INGV (OB.FU.: 1215.010), funded by the Italian Ministry of University and Research. L.G. and E.P. thank the PNRR MUR project PE0000023-NQSTI and Università degli Studi di Catania, project TCMQI PIACERI 2024/2026. E.P. thanks the COST Action CA21144 SuperQumap.

\ifCLASSOPTIONcaptionsoff
  \newpage
\fi



%

\bibliographystyle{IEEEtran}
\bibliography{bibtex/bib/IEEEexample}




%




\end{document}